\documentclass[10pt,twocolumn,letterpaper]{article}

\usepackage[pagenumbers]{cvpr} 

\newcommand{\fst}[1]{\textcolor[rgb]{0,0,0}{\textbf{#1}}}
\newcommand{\snd}[1]{\textcolor[rgb]{0,0,1}{#1}}

\renewcommand{\paragraph}[1]{{\vspace{2pt}\noindent\textbf{#1}}}

\definecolor{cvprblue}{rgb}{0.21,0.49,0.74}
\usepackage[pagebackref,breaklinks,colorlinks,allcolors=cvprblue]{hyperref}

\usepackage{graphicx}
\usepackage{amsmath}
\usepackage{amssymb}
\usepackage{booktabs}
\usepackage{kotex}  
\usepackage{array}
\usepackage{tabularx}
\usepackage{multirow}
\usepackage{subcaption}
\usepackage[dvipsnames]{xcolor}
\usepackage{breakcites} 
\usepackage{comment}
\usepackage{url}
\usepackage{colortbl}
\usepackage{nccmath}
\usepackage{makecell}
\usepackage[accsupp]{axessibility}
\usepackage{afterpage}
\usepackage{siunitx}

\title{Dynamic Exposure Burst Image Restoration}

\author{
Woohyeok Kim ~ ~ ~
Jaesung Rim ~ ~ ~
Daeyeon Kim ~ ~ ~ 
Sunghyun Cho \\ [6mm]
POSTECH
}

\begin{document}
\maketitle
\begin{abstract}
Burst image restoration aims to reconstruct a high-quality image from burst images, which are typically captured using manually designed exposure settings.
Although these exposure settings significantly influence the final restoration performance, the problem of finding optimal exposure settings has been overlooked.
In this paper, we present Dynamic Exposure Burst Image Restoration (DEBIR), a novel burst image restoration pipeline that enhances restoration quality by dynamically predicting exposure times tailored to the shooting environment.
In our pipeline, Burst Auto-Exposure Network (BAENet) estimates the optimal exposure time for each burst image based on a preview image, as well as motion magnitude and gain.
Subsequently, a burst image restoration network reconstructs a high-quality image from burst images captured using these optimal exposure times.
For training, we introduce a differentiable burst simulator and a three-stage training strategy. 
Our experiments demonstrate that our pipeline achieves state-of-the-art restoration quality. 
Furthermore, we validate the effectiveness of our approach on a real-world camera system, demonstrating its practicality. 
\end{abstract}
 
\section{Introduction}
\label{sec:intro}

Burst imaging is a powerful technique in digital photography that involves capturing a series of images in quick succession.
This technique is particularly effective for high-quality image restoration in high-noise scenarios, such as low-light conditions, where high ISO settings are typically used, resulting in significant noise in the output images.
Exploiting the random nature of noise, burst image restoration techniques capture multiple burst images and restore a single high-quality image from them.
Due to its effectiveness, burst image restoration has been actively studied for decades and has become a key feature in modern cameras.

Recent advancements in burst image restoration primarily focus on improving alignment and fusion algorithms~\cite{liba2019handheld,dudhane2022burst,dudhane2023burstormer,nah2019recurrent,zhong2020efficient,bhat2021deep,chan2021basicvsr,chan2022basicvsr++,wu2023rbsr,mehta2023gated}.
Despite these advancements, one crucial aspect of burst imaging has been largely overlooked: finding the optimal exposure settings (i.e., ISO and exposure time) for burst images to maximize restoration quality. Most existing methods rely on manually designed exposure settings, typically assuming uniform exposure settings across all frames. However, this approach results in burst images with the same noise levels and similar degrees of blur, which limits performance due to insufficient complementary information.
Some approaches~\cite{kim2024burst,BracketIRE} have proposed using non-uniform exposure settings across burst images, but they often depend on predefined exposure settings, such as fixed exposure brackets. 
These predefined settings may not always be optimal for varying shooting environments.
For example, when capturing a static scene without camera or object motion, longer exposure times are generally preferred to suppress noise. In contrast, shorter exposure times are more suitable in dynamic scenes with significant motion, as they help mitigate motion blur.

In this paper, we propose a novel burst image restoration pipeline, \emph{Dynamic Exposure Burst Image Restoration (DEBIR)}, which produces a single clean RAW image from burst RAW images in low-light conditions. 
DEBIR enables effective burst image restoration by adaptively predicting an optimal exposure time for each burst image based on the shooting environment. 
To this end, DEBIR consists of a novel \emph{Burst Auto-Exposure Network (BAENet)} and a burst image restoration network. 
BAENet determines the optimal exposure times for burst images, which maximize the restoration network performance, based on a preview image, current exposure settings, and motion information. 
The imaging system then captures burst images using these predicted exposure times, and the restoration network processes them to restore a clean, blur-free, and noise-free image.

However, training BAENet presents a significant challenge, as it necessitates a sequence of ground-truth exposure times for a given scene, which are difficult to obtain. Identifying this ground-truth exposure-time sequence involves comparing the burst image restoration results of burst images with different exposure times for the same scene across all possible combinations.
This task is practically unfeasible because collecting burst images with varying exposure times and their corresponding ground-truth clean images, as well as searching through all possible combinations, is far from straightforward.

To address this challenge, we also propose a novel \emph{differentiable burst simulator} along with an effective three-stage alternating training strategy for BAENet.
Specifically, our differentiable burst simulator takes a set of exposure times as input and synthesizes burst images with realistic degradations, including noise and blur.
The differentiable burst simulator is fully differentiable with respect to exposure times, enabling training of BAENet with respect to the restoration loss.
During training, BAENet predicts the exposure times, and the differentiable burst simulator uses them to synthesize burst images. The burst image restoration network then processes these synthesized burst images to restore a clean image.
We compute a loss between the restored image and the ground-truth clean image and update the weights of BAENet by backpropagating this loss.
This training process eliminates the need for ground-truth exposure times, enabling effective training of BAENet.

We verify the effectiveness of our approach through detailed quantitative and qualitative analyses. 
Our contributions are summarized as follows:
\begin{itemize}
\item[$\bullet$] 
We propose \emph{DEBIR}, a novel burst image restoration pipeline that predicts the optimal exposure times for burst images based on the shooting environment.
To the best of our knowledge, our work is the first to predict optimal exposure time for each burst image by directly optimizing the exposure prediction network with the restoration loss.
\item[$\bullet$]
To effectively train \emph{BAENet}, which predicts the exposure times for burst images, we propose a novel \emph{differentiable burst simulator} along with an efficient training strategy. 
\item[$\bullet$]
We show that \emph{DEBIR} outperforms traditional exposure settings and removes the need for manual optimization, as validated through extensive experiments.
\end{itemize}

\section{Related Work}
\label{sec:relwork}

\paragraph{Burst Imaging.}
Burst imaging has a wide range of applications, including denoising~\cite{dudhane2022burst,dudhane2023burstormer}, deblurring~\cite{nah2019recurrent,zhong2020efficient,Delbracio_2015_CVPR,aittala2018burst}, and super-resolution~\cite{bhat2021deep,chan2021basicvsr,chan2022basicvsr++,wu2023rbsr}. 
Recently, numerous learning-based methods have been proposed.
Bhat~\etal~\cite{bhat2021deep} proposed DBSR that aligns burst images using optical flows and merges them using an attention-based fusion method.
BIPNet~\cite{dudhane2022burst} employs pseudo burst generation and implicit feature alignment using deformable convolution.
Burstormer~\cite{dudhane2023burstormer} adopts a transformer-based architecture for burst imaging.
Mehta~\etal~\cite{mehta2023gated} proposed a multi-resolution architecture and a transposed attention-based fusion method.
However, these approaches only handle burst imaging with uniform exposure and focus on network design, rather than finding the optimal exposure settings.

\paragraph{Restoration using Non-uniform Exposures.}
Capturing multiple images with non-uniform exposures, e.g., using exposure bracketing, has been widely used for HDR imaging~\cite{kalantari2017hdr,yan2019attention,prabhakar2019fast,wu2018deep,niu2021hdrgan,liu2022ghost,yan2023unified,tel2023alignment,zhang2024selfhdr,song2022selective}. 
Beyond HDR imaging, several methods have explored non-uniform exposures for other tasks.
One common strategy is dual-exposure imaging, which captures a pair of long- and short-exposure images and leverages their complementary information for denoising and deblurring~\cite{yuan2007image,chang2021low,mustaniemi2020lsd2,zhao2022d2hnet,zhang2022selfsupervised,shekarforoush2023dual,lai2022face}.
Separately, Kim~\etal~\cite{kim2024burst} recently proposed a method to select the base frame among burst images captured with non-uniform exposures, while Zhang~\etal~\cite{BracketIRE} introduced an exposure bracketing approach for various restoration tasks.
However, these non-uniform exposure approaches rely on fixed brackets, which may lead to suboptimal performance under varying scene dynamics and lighting conditions.

\paragraph{Auto-Exposure (AE).}
Most AE methods aim to adjust exposures to acquire images with proper brightness without considering subsequent image restoration processes~\cite{cameraProduct,shim2018gradient,Su2015Fast,Su2016Model,battiato2010image,schulz2007using,torres2015optimal,park2009method}.
A few studies have explored learning-based AE for specific tasks, such as personalized imaging~\cite{yang2018personalized} and object detection~\cite{onzon2021neural}, yet they typically assume single-image scenarios.
Recently, Xu~\etal~\cite{xu2025adaptiveae} proposed a reinforcement learning-based HDR imaging method that sequentially predicts exposure settings for capturing input LDR images.
However, this stepwise process lengthens the interval between images, making it time-consuming and unsuitable for burst imaging.

For restoration, approaches such as Digital-Gimbal~\cite{dahary2021digital}, Active S-L~\cite{yang2022active}, and Liba~\etal's method~\cite{liba2019handheld} have been proposed. 
Digital-Gimbal introduces learnable exposure parameters for burst images, which are optimized during training, but the parameters remain fixed after training regardless of the scene.
Active S-L treats exposure prediction as a classification problem, selecting the exposure settings from a predefined set.
However, this approach is not scalable, as the predefined set grows exponentially with the number of input images.
Liba~\etal employ motion metering to predict exposure times for burst imaging, but their approach assumes uniform exposure times.
Unlike previous approaches, our method adaptively predicts exposure time for each burst image according to the shooting environment and seamlessly scales to an arbitrary number of images.

\begin{figure*}[t!]
\centering
\includegraphics[width=0.95\linewidth]{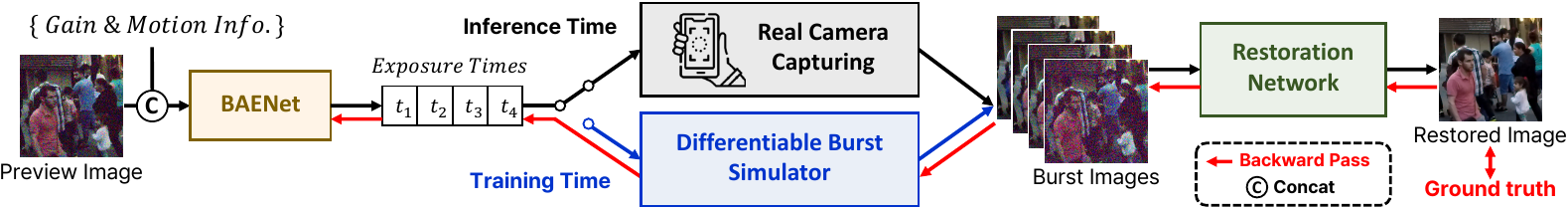}
\vspace{-0.2cm}
\caption{
Overview of our pipeline. BAENet predicts the exposure times of each burst image from a preview image. Differentiable Burst Simulator generates burst images according to the exposure times. The restoration network then reconstructs a high-quality image from them. During inference, the simulator is removed, and the restoration network processes real burst images captured by our camera system.
}
\vspace{-0.4cm}
\label{fig:pipeline}
\end{figure*}

\section{Image Degradation Model}
\label{sec:uniform_vs_nonuniform}

The exposure $e$ of an image $I$ is determined by its exposure time $t$ and gain $g$, as follows:
\begin{equation}
    e=\frac{t\cdot g}{k},
    \label{eq:exposure_time_gain}
\end{equation}
where $k$ is a constant that depends on the specific camera.
The exposure $e$ determines the overall brightness of the image $I$, while the exposure time $t$ and the gain $g$ influence the level of blur and noise, respectively.
Specifically, the effects of $t$ and $g$ on the image $I$ in the camera RAW space can be described by an image degradation model defined as: 
\begin{equation}
I=\mathrm{quant}\circ\mathrm{clip}\circ\mathrm{cfa}\bigl(
  g\bigl(\int_{0}^{t} S_{\tau}\,d\tau + N_{\mathrm{shot}} + N_{\mathrm{read}}\bigr)
\bigr),
\label{eq:image_model}
\end{equation}
where $S_{\tau}$ is the scene radiance at time $\tau$ in the camera RAW color space.
$N_\textrm{shot}$ and $N_\textrm{read}$ are shot noise and read noise, respectively. 
$\textrm{quant}(\cdot)$, $\textrm{clip}(\cdot)$, and $\textrm{cfa}(\cdot)$ represent quantization, dynamic-range clipping, and color filter array (CFA) sampling, respectively.
$\circ$ indicates function composition.
A long exposure time increases the signal-to-noise ratio (SNR), but may also cause blur in $I$ as scene radiance changes over time. 
On the other hand, increasing the gain with short exposure time prevents blur, but amplifies noise.

Most existing burst image restoration approaches capture burst images with the same exposure time and gain. Consequently, the input burst images exhibit similar levels of noise and blur, making it challenging to leverage complementary information for effective burst image restoration.
In contrast, DEBIR captures burst images with non-uniform exposure times and gains to effectively exploit complementary information from them.
Furthermore, BAENet adaptively predicts exposure times according to the shooting environment, enabling more effective burst image restoration.

\section{Dynamic Exposure Burst Image Restoration}
\label{sec:method}

\cref{fig:pipeline} shows an overview of DEBIR.
DEBIR assumes an imaging scenario in which live-streaming preview images are captured and displayed to the user prior to taking burst shots, a common feature in modern cameras.
During the capture of preview images, the imaging system also performs AE to find a proper exposure $e$ for the target scene.

Once the user presses the shutter button, BAENet predicts the optimal exposure times for $n$ burst images, leveraging information derived from preview images.
Specifically, for $n$ burst images $\mathbf{I}=\{I_1, \cdots,I_n\}$, BAENet predicts their optimal exposure times $\mathbf{t}=\{t_1,\cdots,t_n\}$.
The gains of the burst images $\mathbf{g}=\{g_1,\cdots,g_n\}$ are then determined as $g_i = ke/t_i$ where $e$ is the exposure value of the target scene estimated by AE during the preview stage.
As a result, the burst images have the same brightness level but different levels of noise and blur to exploit the complementary information from them.
Subsequently, our framework captures burst images in the camera RAW space using the predicted exposure times and gains, and processes them with a burst image restoration network to produce a clean RAW image.
During training, we simulate this process using a differentiable burst simulator in place of a real imaging system, enabling end-to-end optimization.

Our framework targets low-light conditions, where burst image restoration is most critical.
Based on this, we assume that preview images have a constant exposure time, set to the maximum permissible exposure time for preview, while their gains vary depending on the light condition. In our experiments, we set the exposure time of preview images, $t_p$, to $1/120$ seconds.
In the following, we describe each component of DEBIR in detail.

\subsection{BAENet}
To predict optimal exposure times reflecting the shooting environment, BAENet takes a preview image $I_p$ in the camera RAW space, its gain $g_p$, and motion magnitude $m_p$ as inputs.
We define the motion magnitude $m_p$ as the average magnitude of the optical flow vectors estimated between $I_p$ and its previous preview frame $I'_p$.
For optical flow estimation, we convert the RAW $I_p$ and $I'_p$ to sRGB images by applying a simple ISP pipeline~\cite{brooks2019unprocessing}, and adopt an off-the-shelf optical flow estimator~\cite{teed2020raft}.
From $I_p$, $g_p$, and $m_p$, BAENet predicts $n$ exposure times $\mathbf{t}$, from which their corresponding gains $\mathbf{g}$ are computed using \cref{eq:exposure_time_gain}.

By leveraging the gain $g_p$ and the motion magnitude $m_p$, BAENet adaptively predicts optimal exposure times based on the amount of noise, and on both camera motion and object motion in the target scene.
Additionally, $I_p$ enables BAENet to more accurately reflect the characteristics of the target scene, such as detailed distributions of noise and blur, which are not captured by $g_p$ and $m_p$, as well as its content.

For the network architecture of BAENet, we adopt MobileNetV2~\cite{sandler2018mobilenetv2} due to its lightweight design and efficient computation.
Specifically, for the input to BAENet, we construct a tensor by concatenating $g_p$ and $m_p$ to $I_p$ along the channel dimension, and feed this tensor to BAENet.
We modify the input channel size of MobileNetV2 accordingly. 
As $g_p$ and $m_p$ have large values on different scales, we normalize them so that they are in the range of $[0,1]$ by simple shifting and scaling. We refer the reader to the supplementary material for more implementation details on BAENet.

Training BAENet to predict arbitrarily long exposure times is challenging and unstable due to the unbounded search space.
Thus, for stable training, we constrain the sum of the exposure times to be shorter than a predefined upper bound $t_\textrm{u}$, i.e., $\sum_i t_i \leq t_\textrm{u}$.
To achieve this, we modify the last layer of MobileNetV2 to produce a vector of $n+1$ dimensions and apply the bounded softmax function, ensuring that the sum of the output is 1, and that each element is positive. 
We then multiply the first $n$ elements of the output vector by $t_\textrm{u}$ to obtain the exposure times of the burst images.
The $(n+1)$-th element of the output vector allows the sum of the $n$ exposure times to be shorter than $t_\textrm{u}$, if necessary.
Mathematically, $t_i$ is obtained as:
\begin{equation}
    t_i = t_\textrm{u} \cdot \textrm{softmax}_\textrm{bounded}(f_i,\epsilon),
\end{equation}
where $f_i$ is the $i$-th feature value in the last layer before the bounded softmax operation, and $\textrm{softmax}_\textrm{bounded}(f_i,\epsilon)$ is the bounded softmax function of $f_i$ whose output is bounded into $[\epsilon,1-n\epsilon]$. 
We set $\epsilon = t_\textrm{min}/t_\textrm{u}$ where $t_\textrm{min}$ is the minimum exposure time of our imaging system ($1/240$ sec.).

\subsection{Burst Image Restoration Network}
\label{ssec:burst_image_restoration_network}

Once BAENet predicts optimal exposures for burst images, our imaging system captures burst images where each burst image $I_i$ is a RAW image with the Bayer pattern.
The burst image restoration network takes $n$ RAW images as input and produces a single restored RAW image.
For the architecture of the restoration network, we employ Burstormer~\cite{dudhane2023burstormer}, a state-of-the-art model for burst image restoration and enhancement tasks.
Specifically, we adopt a variant of the Burstormer architecture modified for burst denoising\footnote{\url{https://github.com/akshaydudhane16/Burstormer}}.
We adjust the channel sizes of the input layer and of the intermediate convolution layers, which are determined by the number of input images.

\subsection{Differentiable Burst Simulator}
\label{ssec:DBS}

To synthesize burst images with varying exposures and preview images while enabling end-to-end training of BAENet with the restoration loss, our training process employs a differentiable burst simulator, which is a module with no learnable parameters.
Given exposure times $\mathbf{t}$, gains $\mathbf{g}$, and a scene radiance sequence $\mathbf{S}$, which will be defined later, the simulator generates burst images. To this end, it first computes the start time $t^s_i$ and end time $t^e_i$ of the exposure duration for each burst image $I_i$. Specifically, we calculate $t^s_i$ and $t^e_i$ as:
\begin{equation}
    t_1^s=t^{0},~~t^s_i = t^e_{i-1} + \delta~~\textrm{for}~~i>1,~~\textrm{and}~~t^e_i=t^s_i+t_i,
\end{equation}
where $t^0$ is a constant to reserve a certain amount of time for preview images before the first burst image.
$\delta$ represents the inter-frame gap between consecutive burst images, which we set $\delta=7/1920$ sec.~in our experiments.

Next, the simulator synthesizes each degraded RAW image $I_i$ with realistic blur and noise corresponding to its exposure duration $[t_i^s, t_i^e]$ and gain $g_i$. Mathematically, the synthesis of each burst image $I_i$ can be described as:
\begin{equation}
    I_i = \textrm{syn}(\mathbf{S}, t^s_i, t^e_i, g_i),
\end{equation}
where $\textrm{syn}(\cdot)$ represents a degraded image synthesis function, which is defined based on the image degradation model in \cref{eq:image_model}.
The scene radiance sequence $\mathbf{S}$ models the varying scene radiance $S_\tau$ in \cref{eq:image_model} and is defined as a high-FPS video frame sequence in the camera RAW color space, i.e., $\mathbf{S} = \{S_0, S_1, \cdots\}$. Each $S_i$ in $\mathbf{S}$ indicates scene radiance during a small exposure time $e_S$, which is set to $1/1920$ sec.~in our experiments.

To support back-propagation through the simulator, it must be differentiable with respect to the input exposure times $\mathbf{t}$. To achieve this, the degraded image synthesis function is designed to be differentiable with respect to $t^s_i$ and $t^e_i$, supporting continuous values for $t^s_i$ and $t^e_i$. Specifically, we define the degraded image synthesis function as:
\begin{equation}
    \textrm{syn}(\mathbf{S}, t^s, t^e, g) = \textrm{clip}\circ\textrm{cfa}\left( S_{s,e} + gN \right) \label{eq:degraded_image_synthesis_1},
\end{equation}
where $S_{s,e}$ represents the integration of scene radiance during the exposure duration from $t^s$ to $t^e$, and $N$ models both photon shot noise and read noise.
We define $S_{s,e}$ as:
\begin{equation}
    S_{s,e} = \frac{1}{\bar{t}^e - \bar{t}^s} \left( \alpha_s S_{\lfloor \bar{t}^s \rfloor} + \sum_{\tau = \lceil \bar{t}^s \rceil}^{\lfloor \bar{t}^e \rfloor} S_{\tau} + \alpha_e S_{\lceil \bar{t}^e \rceil} \right),
\end{equation}
where $\bar{t}^s=t^s/e_S$ and $\bar{t}^e=t^e/e_S$. $\alpha_s$ and $\alpha_e$ are blending weights for the first and last scene radiance frames defined as $\alpha_s = \lceil \bar{t}^s \rceil - \bar{t}^s$ and $\alpha_e = \bar{t}^e - \lfloor \bar{t}^e \rfloor$, respectively.
These weights ensure smooth interpolation across frames, making $S_{s,e}$ differentiable with respect to the exposure times $\mathbf{t}$.
For the noise $N$ in \cref{eq:degraded_image_synthesis_1}, we model both photon shot noise and read noise using a heteroscedastic Gaussian distribution, following \cite{BracketIRE}: 
\begin{equation}
    N \sim \mathcal{N}\left(0, \lambda_\textrm{read} + \lambda_\textrm{shot} S_{s,e} \right)
    \label{eq:noise_sampling},
\end{equation}
where $\lambda_\textrm{shot}$ and $\lambda_\textrm{read}$ correspond to the shot noise and read noise parameters, respectively.
However, directly implementing \cref{eq:noise_sampling} is non-differentiable. 
To circumvent this, we adopt the reparameterization trick~\cite{kingma2014auto,rezende2014stochastic}.
Specifically, we reformulate \cref{eq:noise_sampling} as $N = \sqrt{\lambda_{\text{read}} + \lambda_{\text{shot}} S_{s,e}} \cdot Z$, where $Z \sim \mathcal{N}(0, 1)$.
This formulation renders the random variable $N$ differentiable with respect to $S_{s,e}$, and ultimately with respect to $\mathbf{t}$.
Details of the gain range and noise parameters are provided in the supplementary material.

Compared to \cref{eq:image_model}, the degraded image synthesis function in \cref{eq:degraded_image_synthesis_1} has two differences. Firstly, it does not include quantization to ensure non-zero gradients during training.
Secondly, it does not amplify scene radiance but amplifies only the noise $N$ according to the gain $g$.
This is because our framework assumes that input burst images have gains inversely proportional to their exposure times, which results in the same brightness level across the images while producing different noise levels.

Training of our framework requires not only burst images but also their corresponding ground-truth images and preview images.
For the ground-truth image $I_{gt}$, we use the first sharp frame of $I_1$, i.e., $I_{gt}= \textrm{cfa}(S_{\bar{t}^0})$.
Regarding preview images $I_p$ and $I'_p$, we synthesize them using the differentiable burst simulator as follows.
First, we set the exposure durations $[t_p^s,t_p^e]$ and $[t_p'^s,t_p'^e]$ of $I_p$ and $I'_p$ as:
\begin{align} 
  t_p^e=t^0-\delta,~~~~t_p^s=t_p^e-t_p,\\t_p'^e=t_p^s-\delta_p,~~~~t_p'^s=t_p'^e-t_p,   
\end{align}
where $\delta_p$ represents the temporal gap between the two preview images, which we set $39/1920$ seconds.
We set $t^0$ to be a multiple of $e_S$ that is greater than $2t_p + \delta_p + \delta$ to take into account the exposure times of the preview images and the temporal gaps, and to ensure that $I_{gt}$ has sharp details and aligns well with $I_1$.
Then, using the differentiable burst simulator, we synthesize $I_p$ and $I'_p$ with their respective exposure durations $[t_p^s,t_p^e]$ and $[t_p'^s,t_p'^e]$ and the gain $g_p$, which is randomly sampled during training.

\section{Training}
\label{ssec:training}

While our differentiable burst simulator enables end-to-end training of both BAENet and the burst image restoration network, jointly training them from scratch is unstable and prone to local minima due to their mutual dependence.
In early training, BAENet may initially predict random suboptimal exposure times, which can cause the restoration network to adapt prematurely to these exposure times. This early bias in the restoration network can then propagate back to BAENet, reinforcing exposure predictions that align with the network’s limited capabilities rather than the true optimal ones.
To mitigate this issue, we adopt a three-stage strategy: (1) we first pre-train the restoration network to handle burst images with diverse exposure times, (2) we then train BAENet to predict the optimal exposure times using the pre-trained restoration network, and (3) we finally fine-tune the restoration network with BAENet to further enhance performance.
In the following, we describe our training dataset and each stage in detail.

\subsection{Datasets}
To train our framework, we use a training dataset $\mathcal{D}$ consisting of scene radiance sequences, i.e., $\mathcal{D} = \{\mathbf{S}_1,\mathbf{S}_2,\cdots\}$. 
We synthesized our scene radiance dataset from a video dataset consisting of clean video clips.
Specifically, for each video clip in the video dataset, we 
apply gamma expansion to convert them into the linear sRGB space. 
Afterward, we convert the videos into the RAW color space by randomly sampling inverse color correction matrices (CCMs) and inverse white balance (WB) gains, and applying them to the videos.
For the video dataset, we used video clips from the GoPro dataset~\cite{nah2017deep}, where each video clip is captured at 240 FPS and contains camera and object motions. By applying frame interpolation, we increase the frame rate by a factor of 8, resulting in video clips with a frame rate of 1920 FPS.
Additionally, we also use the RealBlur dataset~\cite{rim_2020_ECCV} to synthesize scene radiance sequences of static scenes. 
Additional dataset details are provided in the supplementary material.
Through this process, we generated 5,219 scene radiance sequences for training and 532 for evaluation.

To prevent overfitting across training stages, we avoid reusing the same sequences for both the burst image restoration network and BAENet.
Without this precaution, the burst image restoration network might become biased toward certain combinations of burst images and exposure times encountered during the first training stage. 
This bias could adversely affect BAENet in the second stage, causing it to favor those specific combinations and ultimately impairing its generalization capability. 
Thus, to mitigate this issue, we split the dataset $\mathcal{D}$ into $\mathcal{D}_\textrm{restore}$ (4,092 sequences) and $\mathcal{D}_\textrm{BAENet}$ (1,127 sequences).

\subsection{Pre-training Burst Image Restoration Network}
\label{ssec:training_burstormer}

Our training process starts with training the burst image restoration network.
In this step, we train the restoration network using a loss $\mathcal{L}_\textrm{restore}$ defined as:
\begin{equation}
    \mathcal{L}_\textrm{restore} = \| \textrm{res}_\phi(\mathbf{I})-I_\textrm{gt}\|_1,
\end{equation}
where $\textrm{res}_\phi(\cdot)$ is the restoration network parameterized by $\phi$. $\mathbf{I}$ is a burst image sequence synthesized using our differentiable burst simulator with randomly sampled exposure times and gains, and $I_\textrm{gt}$ is the ground-truth image.

We synthesize the burst images using the following method.
We first randomly sample a scene radiance sequence $\mathbf{S}$ from $\mathcal{D}_\textrm{restore}$, and a gain $g_p$ for the sampled sequence.
Next, we randomly sample the exposure time $t_i$ for the $i$-th burst image from $[t_\textrm{min}, t_\textrm{max}]$, where $t_\textrm{min}$ and $t_\textrm{max}$ are set to $1/240$ sec. and $1/30$ sec., respectively.
We then set its gain $g_i$ as $g_i = t_p \cdot g_p / t_i$, ensuring that all burst images in the sequence have the same brightness. Finally, using the sampled scene radiance sequence $\mathbf{S}$, and the sampled exposure times $t_i$ and gains $g_i$, we synthesize burst images with the differentiable burst simulator.

\subsection{Training BAENet}
Once the burst image restoration network is trained, we train BAENet using the restoration network on $\mathcal{D}_\textrm{BAENet}$.
We train BAENet in two steps, a warm-up step and a main training step, to avoid local minima and to achieve high-quality exposure time predictions.

The warm-up step trains BAENet using a loss $\mathcal{L}_\textrm{warm-up}$:
\begin{equation}
    \mathcal{L}_\textrm{warm-up}=\|\textrm{bae}_\theta(I_p,g_p,m_p)-\mathbf{t}_\textrm{pseudo-gt}\|_1,
\end{equation}
where $\textrm{bae}_\theta(\cdot)$ represents BAENet parameterized by $\theta$, and $\mathbf{t}_\textrm{pseudo-gt}$ is a vector with pseudo-ground-truth exposure times obtained by examining a predefined set of different combinations of exposure times.
For the warm-up step, we generate a training dataset with pseudo-ground-truth exposure times as follows.
First, we define a set $\mathcal{E}=\{\hat{\mathbf{t}}_1,\cdots\}$ where $\hat{\mathbf{t}}_i$ is a predefined combination of exposure times.
In our experiments, we define $\mathcal{E}$ as $\mathcal{E}=\{(8,8,8,8),$ $(16,16,16,16),$ $(24,24,24,24),$ $(32,32,32,32),$ $(8,16,24,32)\}/1920$ seconds.
Next, we sample a scene radiance sequence $\mathbf{S}$ from $\mathcal{D}_\textrm{BAENet}$ and synthesize the input to BAENet, i.e., a gain $g_p$, preview images $I_p$ and $I'_p$, and motion magnitude $m_p$, as well as the ground-truth image $I_{gt}$.
We then synthesize burst images for each of the exposure time combinations in $\mathcal{E}$, and perform burst image restoration to each burst image sequence, and compare their results to $I_{gt}$.
Finally, we choose the exposure time combination of the best result as $\mathbf{t}_\textrm{pseudo-gt}$.
Once the dataset is generated, we train BAENet using $\mathcal{L}_\textrm{warm-up}$.

After the warm-up step, the main training step trains BAENet using a loss $\mathcal{L}_\textrm{DEBIR}$:
\begin{equation}
    \mathcal{L}_\textrm{DEBIR} = \|\textrm{res}_\phi(\textrm{sim}(\textrm{bae}_\theta(I_p,g_p,m_p)))-I_\textrm{gt}\|_1,
    \label{eq:loss_DEBIR}
\end{equation}
where $\textrm{sim}(\cdot)$ represents the differentiable burst simulator.
In this step, we fix the restoration network, update only BAENet for stable training of BAENet.

\subsection{Fine-tuning Burst Image Restoration Network}

Once BAENet is trained, we fine-tune the burst image restoration network to further improve its performance for the predicted exposure times by BAENet.
To this end, we minimize $\mathcal{L}_\textrm{DEBIR}$ in \cref{eq:loss_DEBIR} with respect to the parameters of the restoration network for $\mathcal{D}_\textrm{restore}$, while fixing BAENet.

\section{Experiments}
\label{sec:experiments}

\paragraph{Implementation}
We implemented DEBIR using PyTorch~\cite{paszke2017automatic}.
We pre-trained the burst image restoration network for 500 epochs, trained BAENet for 100 epochs, and fine-tuned the burst network for 50 epochs.
The learning rates for each stage were set to \num{3e-4}, \num{1e-7}, and \num{1e-5}, respectively, and were decayed to \num{1e-8}, \num{1e-8}, and \num{1e-7} using a cosine annealing scheduler~\cite{loshchilov2017sgdr}. 
We used the AdamW optimizer~\cite{loshchilov2017decoupled}, and stage~2 included a 35-epoch warm-up training.
All experiments, except for the ablation on the number of burst images, were conducted with an upper bound of exposure time $t_\textrm{u}=128/1920$ sec. and a burst size of $n=4$, using training images of resolution $256\times256$.
Training was conducted on a PC equipped with four GeForce RTX 3090 GPUs, using a batch size of 4.

\begin{figure*}[t!]
\centering
\includegraphics[width=0.9\linewidth]{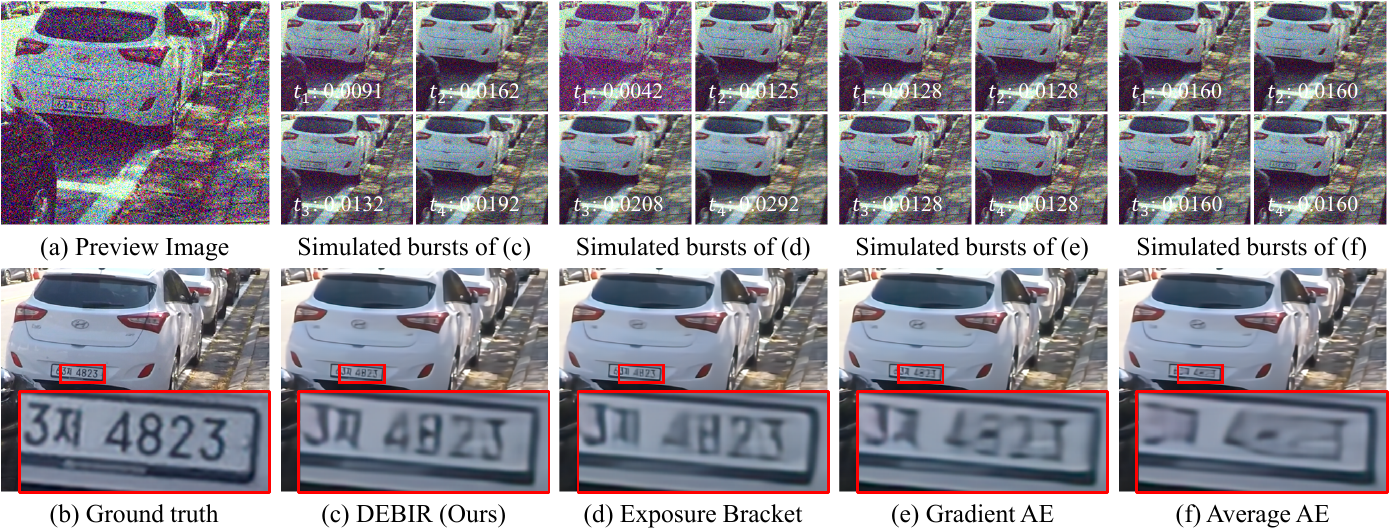}
\vspace{-0.2cm}
\caption{Qualitative results on our test set. See the supplementary material for more results.
}
\vspace{-0.4cm}
\label{fig:qual_syn}
\end{figure*}

\subsection{Comparison with Other Methods}
\label{ssec:bir}

We compare DEBIR with existing AE and predefined-exposure methods. 
Each method either predicts exposure times or uses predefined ones, and the differentiable burst simulator generates the corresponding burst images, which are then restored by the restoration network.
We evaluate classical and learning-based AE methods. 
The classical baselines, Average AE~\cite{cameraProduct} and Gradient AE~\cite{shim2018gradient}, determine exposure based on global brightness and local gradients.
Since they produce a single exposure time, we apply the exposure to all burst images. 
For learning-based methods, we include Digital-Gimbal~\cite{dahary2021digital} and modify it by introducing learnable exposure parameters for each burst image.
We also modify Active S-L~\cite{yang2022active}, originally a two-image gain classifier, to estimate exposure times instead of gains.
As a predefined-exposure baseline, we adopt classical exposure bracketing~\cite{kim2024burst,BracketIRE} with increasing exposure times of $\{8,24,40,56\}/1920$ seconds.

\begin{table}[t!]
    \centering
    \scalebox{0.88}{
    \begin{tabularx}{1.03\columnwidth}{c|ccc}
    \Xhline{2.5\arrayrulewidth}
    Methods & PSNR$\uparrow$ & SSIM$\uparrow$ & LPIPS$\downarrow$ \\ \hline
    Digital-Gimbal~\cite{dahary2021digital} & 33.87 & 0.9309 & 0.187  \\ 
    Active S-L~\cite{yang2022active} & 33.89 & 0.9379 & 0.176 \\
    Average AE~\cite{cameraProduct} & 34.69 & 0.9484 & 0.157 \\
    Gradient AE~\cite{shim2018gradient} & 34.86 & \snd{0.9494} & \snd{0.156} \\
    Exposure Bracket~\cite{kim2024burst,BracketIRE} & \snd{35.04} & 0.9481 & 0.164 \\ 
    DEBIR (Ours) & \fst{35.32} & \fst{0.9519} & \fst{0.154} \\
    \Xhline{2.5\arrayrulewidth}
    \end{tabularx}
    }
    \vspace{-0.2cm}
    \caption{Quantitative comparisons on our test set. The best and second-best results are in bold and blue, respectively.}
\vspace{-0.4cm}
    \label{tab:comparison}
\end{table}

All methods are integrated with Burstormer~\cite{dudhane2023burstormer} and are trained on our training set.
For Active S-L, we integrated it with Burstormer for two images, as the method is not scalable beyond two images. 
To ensure fairness, we applied the same pre-training and fine-tuning steps to the restoration network for all methods.

\cref{tab:comparison} presents quantitative results, where DEBIR outperforms all baselines.
Average AE and Gradient AE predict a single exposure time that is applied to all burst images, leading to similar noise and blur levels in the burst images, which ultimately limits their restoration performance.
Predefined exposure bracketing performs better since it uses non-uniform exposures, but its fixed schedule still cannot adapt to the scene, leading to suboptimal results.
Among learning-based methods, Digital-Gimbal is limited because its exposure parameters are fixed after training, while Active S-L underperforms due to its two-image design that cannot scale to longer bursts, both leading to insufficient performance gains.
In contrast, DEBIR predicts per-frame exposure times conditioned on the shooting environment, better balancing noise and blur and exploiting stronger complementary information.
\cref{fig:qual_syn} shows that DEBIR generates the most visually pleasing results.

\subsection{Analysis of BAENet}

\begin{figure}[t!]
\includegraphics[width=0.9\linewidth]{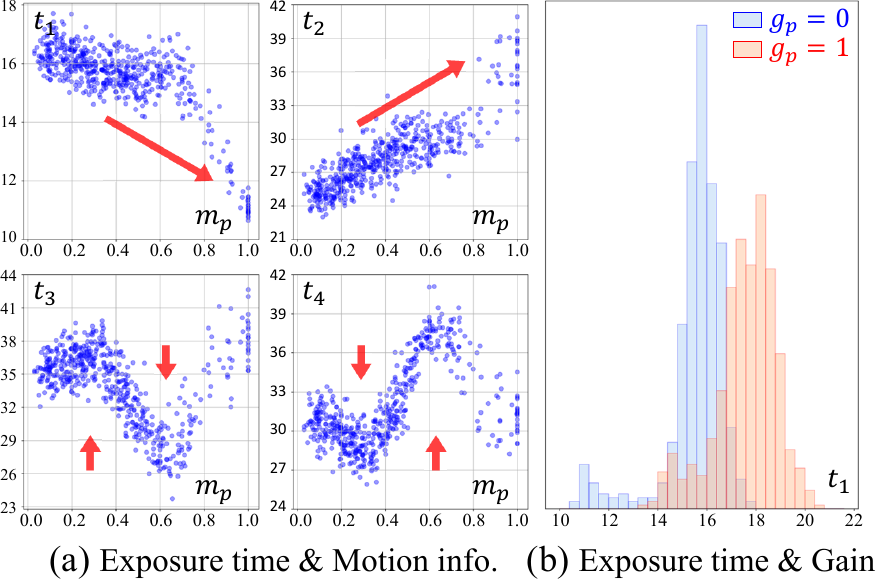}
\centering
\vspace{-0.25cm}
\caption{Analysis of predicted exposure times. (a) Scatter plots of exposure times of each frame $t_i$ and motion magnitude $m_p$. 
(b) Exposure time histograms of the first frame ($t_1$) for the minimum and maximum preview image gains $g_p$. 
The unit of exposure times is $1/1920$ sec., and $g_p$ is normalized to $[0,1]$.
}
\vspace{-0.4cm}
\label{fig:analysis}
\end{figure}

\paragraph{Predicted Exposure Time} 
We analyze the predicted exposure time of each burst image $t_i$ in relation to motion magnitude $m_p$ and preview image gain $g_p$.
\cref{fig:analysis}-(a) and (b) show scatter plots of $t_i$ versus $m_p$ at $g_p=0$ and histograms of $t_1$ at the minimum and the maximum $g_p$, respectively.
The plots reveal two key observations. 
First, the exposure time of the first frame $t_1$ decreases as $m_p$ increases to mitigate blur (\cref{fig:analysis}-(a)).
Additionally, increasing the gain $g_p$, which implies higher-noise conditions, shifts the predicted $t_1$ toward longer exposure times to suppress noise (\cref{fig:analysis}-(b)).
Second, BAENet optimizes the exposure times to maximize complementary information from adjacent burst images. 
For instance, in high-motion scenarios where $m_p$ is large, BAENet decreases $t_1$ to minimize blur in the first frame. However, this also amplifies noise.
To compensate for this, BAENet increases $t_2$ to capture complementary information, such as color details and noise-free information, which help mitigate noise in the first frame.
A similar trend is also found between the third and fourth frames, where $t_4$ adjusts in the opposite direction of $t_3$ (\eg, red arrows in \cref{fig:analysis}-(a)). 
This result demonstrates the importance of complementary information in burst image restoration. 

\begin{figure}[t!]
\includegraphics[width=1.0\linewidth]{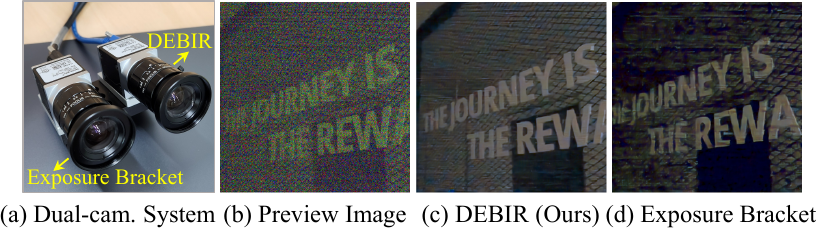}
\centering
\vspace{-0.6cm}
\caption{Qualitative results using a real-world camera system.}
\vspace{-0.2cm}
\label{fig:real}
\end{figure}

\begin{table}[t!]
    \centering
    \scalebox{0.88}{
    \begin{tabularx}{1.125\columnwidth}{c|c@{\hspace{0.45em}}c@{\hspace{0.45em}}c}
    \Xhline{2.5\arrayrulewidth}
    Methods & NIQE$\downarrow$~\cite{mittal2012making} & BRISQUE$\downarrow$~\cite{mittal2012no} & TOPIQ$\uparrow$~\cite{chen2023topiq} \\ \hline
    Exposure Bracket & 6.57 & 50.66 & 0.339 \\ 
    DEBIR (Ours) & 6.34 & 46.90 & 0.363 \\
    \Xhline{2.5\arrayrulewidth}
    \end{tabularx}
    }
    \vspace{-0.2cm}
    \caption{Quantitative results using a real-world camera system.}
\vspace{-0.4cm}
    \label{tab:real}
\end{table}

\paragraph{Real-world Capturing and Evaluation} 
We capture real-world burst images using exposures predicted by BAENet, and compare the restored images to those from predefined-exposure settings~\cite{kim2024burst,BracketIRE}.
For a fair comparison, burst images from both methods must be captured simultaneously, as scene motion and content can change.
To this end, we construct a handheld dual-camera system, depicted in \cref{fig:real}-(a).
The system consists of two cameras: one captures burst images using BAENet, while the other simultaneously captures burst images using exposure bracketing.
We used the models trained on our dataset (\cref{tab:comparison}). 
The inference time of BAENet, including optical flow estimation, is 0.023 seconds.
Using the camera system, we captured and evaluated 142 burst images in low-light environments. 

As ground-truth images are unavailable in practice, we evaluate the restored images using no-reference metrics.
As shown in \cref{fig:real}, the exposure bracket fails to restore accurate colors and produces artifacts. 
In contrast, our method successfully restores a high-quality image. \cref{tab:real} further validates the effectiveness of our method in real-world scenarios, highlighting its superior performance over predefined exposure settings.
Additional results and details of our camera system are provided in the supplementary material.

\subsection{Ablation Study}
\label{ssec:ablation}

\begin{table}[t]
    \centering
    \scalebox{0.88}{
    \begin{tabularx}{1.12\columnwidth}{c@{\hspace{0.9em}}c@{\hspace{0.7em}}c@{\hspace{0.7em}}c@{\hspace{0.7em}}c}
    \Xhline{2.5\arrayrulewidth}
    Inputs & w/o Prev. Image & w/o Gain & w/o Motion Info. & Full \\ \hline
    PSNR$\uparrow$ & 34.80 & \snd{35.21} & 35.13 & \fst{35.32} \\  
    \Xhline{2.5\arrayrulewidth}
    \end{tabularx}
    }
    \vspace{-0.2cm}
    \caption{Ablation study on the impact of each input in BAENet. The best and second-best results are in bold and blue, respectively.}
    \vspace{-0.2cm}
    \label{tab:abl-baenet}
\end{table}

\paragraph{Uniform vs. Non-uniform Exposures}
In this study, we examine the impact of the non-uniform exposure setting in our method.
To this end, we modify BAENet to predict a single exposure time and apply it uniformly across the burst.
We then train the modified BAENet and the burst image restoration network using our training strategy.
The variant achieves 35.01 dB PSNR, which is 0.31 dB lower than our final model.
This result again shows the advantage of non-uniform exposure settings over uniform ones.

\paragraph{Inputs of BAENet} 
BAENet takes a preview image, gain, and motion magnitude as inputs. 
We examine the impact of each input component.
\cref{tab:abl-baenet} shows that excluding any input significantly reduces performance, indicating that all inputs are crucial.
Among them, the preview image has the most significant impact, as it contains all relevant information.
However, providing motion magnitude and gain information helps quantify motion levels and noise, improving performance.
Also, the table shows that the impact of motion information is more significant than that of gain.

\begin{table}[t!]
    \centering
    \scalebox{0.88}{
    \begin{tabularx}{1.07\columnwidth}{ccccc}
    \Xhline{2.5\arrayrulewidth}
    \# of burst images & 2 & 4 & 6 & 8 \\ \hline
    Exposure Bracket~\cite{kim2024burst,BracketIRE} & 33.59 & 35.04 & 35.69 & \snd{35.89} \\
    DEBIR (Ours) & 34.22 & 35.32 & 35.84 & \fst{36.11} \\
    \Xhline{2.5\arrayrulewidth}
    \end{tabularx}
    }
    \vspace{-0.2cm}
    \caption{Analysis of the impact of the number of burst images on performance in terms of PSNR. The best and second-best results are in bold and blue, respectively.}
    \vspace{-0.4cm}
    \label{tab:abl-numbursts-compare}
\end{table}

\paragraph{Number of Burst Images}
While our experiments evaluate DEBIR with a fixed burst size of $n=4$, the proposed method can be easily extended to other burst lengths.
To verify this, we train DEBIR with $n = 2$, $4$, $6$, and $8$, and evaluate its performance accordingly.
For comparison, we also evaluate the performance of exposure bracketing methods~\cite{kim2024burst,BracketIRE}. 
In these settings, the total exposure time is scaled proportionally to the burst size (i.e., $64$, $128$, $192$, and $256$ $/1920$ sec.), with the first exposure time fixed at $8/1920$ sec. and subsequent exposure times increasing linearly (e.g., for $n = 4$, $\{8, 24, 40, 56\}/1920$ sec.).
\cref{tab:abl-numbursts-compare} shows that DEBIR outperforms the predefined exposure setting for all burst sizes.
Performance improves with larger burst sizes, as more frames generally provide richer information.

\paragraph{Training Strategy}
We adopt a three-stage training strategy: pre-training the burst image restoration network (S1), training BAENet with $\mathcal{L}_\textrm{warm-up}$ (S2-1), training BAENet with $\mathcal{L}_\textrm{DEBIR}$ (S2-2), and fine-tuning the restoration network (S3).
\cref{tab:strategy} shows the impact of each stage and comparisons with end-to-end training.
The table shows that training BAENet with $\mathcal{L}_\textrm{warm-up}$ (S2-1) improves overall performance by providing pseudo-ground-truth exposure times in the warm-up stage.
After training BAENet, fine-tuning the restoration network (S3) significantly improves the performance. 
This step further optimizes the network to adapt to optimal exposure settings estimated from BAENet.
We also compare our approach with the end-to-end training strategy.
Instead of fine-tuning the restoration network (S3), we apply end-to-end training.
Additionally, we evaluate end-to-end training without the three-stage strategy.
As shown in \cref{tab:strategy}, end-to-end training leads to a significant performance drop, emphasizing the importance of our training strategy.

\begin{table}[t!]
    \centering
    \scalebox{0.88}{
    \begin{tabularx}{0.91\columnwidth}{l|ccc}
    \Xhline{2.5\arrayrulewidth}
    Training Strategy & PSNR$\uparrow$ & SSIM$\uparrow$ & LPIPS$\downarrow$ \\ \hline
    S1, S2-2 & 34.93 & 0.9482 & 0.162 \\
    S1, S2-1, S2-2 & 35.01 & 0.9489 & 0.160 \\ 
    S1, S2-2, S3 & \snd{35.16} & \snd{0.9502} & \snd{0.157} \\ 
    S1, S2-1, S2-2, S3 & \fst{35.32} & \fst{0.9519} & \fst{0.154} \\
    S1, S2-1, E2E & 34.99 & 0.9496 & 0.158 \\
    S1, S2-1, S2-2, E2E & 35.11 & 0.9500 & 0.158 \\
    E2E & 33.60 & 0.9279 & 0.196 \\
    \Xhline{2.5\arrayrulewidth}
    \end{tabularx}
    }
    \vspace{-0.2cm}
    \caption{Ablation study on training strategy. S1: Pre-training the restoration network. S2-1: Training BAENet with $\mathcal{L}_\textrm{warm-up}$. S2-2: Training BAENet with $\mathcal{L}_\textrm{DEBIR}$. S3: Fine-tuning the restoration network. 
    E2E: end-to-end training of BAENet and the restoration network.
    The best and second-best results are in bold and blue.
    }
    \vspace{-0.4cm}
    \label{tab:strategy}
\end{table}

\section{Conclusion}
In this paper, we propose the Dynamic Exposure Burst Image Restoration (DEBIR) pipeline.
Specifically, we introduce BAENet, which estimates exposure times to maximize complementary information across burst images.
Then, a burst restoration network reconstructs high-quality images from burst images captured accordingly.
For training them, we also present a novel Differentiable Burst Simulator and an efficient training strategy.
We validate the effectiveness of DEBIR through extensive experiments.
Finally, we integrate BAENet with a camera system and demonstrate the superior performance of DEBIR in real-world scenarios.

\paragraph{Limitations and Future Work}
BAENet may not always yield the optimal exposure settings for a given scene, as better choices could in principle be found through exhaustive search. 
This limitation is partly due to predicting exposure times for images yet to be captured with limited information and the risk of convergence to local minima. Addressing these issues presents an interesting direction for future work.
Another promising future direction would be to extend our approach to jointly predict both exposure times and burst size, taking into account not only restoration quality but also practical constraints such as power consumption.

\paragraph{Acknowledgments}
This work was supported by grants from the Korea government (MSIT) through the National Research Foundation of Korea (NRF) (No. 2023R1A2C200494611) and the Institute of Information \& Communications Technology Planning \& Evaluation (IITP) (IITP-2026-RS-2024-00437866, No. RS-2019-II191906, Artificial Intelligence Graduate School Program (POSTECH)), as well as by the Basic Science Research Program through the NRF funded by the Ministry of Education (2022R1A6A1A03052954). It was also supported by Samsung Electronics Co., Ltd (IO251210-14286-01).

{
    \small
    \bibliographystyle{ieeenat_fullname}
    \bibliography{references}
}

\end{document}


\maketitle
\thispagestyle{empty}

In this supplementary material, we first present additional details on BAENet (\cref{sec:baenet-supp}), followed by descriptions of the noise synthesis process (\cref{sec:noise}), RAW conversion pipeline (\cref{sec:raw}), and details of the dataset (\cref{sec:dataset}). 
We then describe the camera system for real-world evaluation (\cref{sec:camera}) and computational cost analysis (\cref{sec:cost}). 
Next, we present additional experimental results (\cref{sec:experiments-supp}).
Finally, we include additional qualitative results (\cref{sec:results-supp}).

\section{Implementation Details of BAENet}
\label{sec:baenet-supp}

\paragraph{Normalization of Inputs} 
BAENet takes a preview image along with gain $g_p$ and motion magnitude $m_p$ as inputs.
Since $g_p$ and $m_p$ have different scales with large variations, we normalize them to the range $[0,1]$ as follows: 
\begin{align}
    g_p &= (\hat{g}_p - g_\textrm{min}) / (g_\textrm{max} - g_\textrm{min}) \in [0,1], \\
    & \;\; m_p = \min{(\hat{m}_{p}/m_\textrm{thr},1)} \in [0,1],
\end{align}
where $\hat{g}_p$ and $\hat{m}_{p}$ are the gain and motion magnitude before the normalization, respectively. 
$g_\textrm{min}$ and $g_\textrm{max}$ represent the minimum and maximum gain values for the shooting environment, respectively. We set $g_\textrm{min} = 51200$ and $g_\textrm{max} = 102400$ to account for extremely low-light conditions.
$m_\textrm{thr}$ denotes the threshold of motion magnitude and is set to 20. We empirically found that truncating motion values beyond this threshold improves performance.

\paragraph{Network Architecture}
For BAENet, we adopt MobileNetV2~\cite{sandler2018mobilenetv2}, and construct the input by concatenating  $g_p$ and $m_p$ to each pixel position of $I_p\in\mathbb{R}^{H \times W \times 3}$, then feed the concatenated tensor $B_\textrm{input}\in\mathbb{R}^{H \times W \times 5}$  into BAENet.
We modify the input channels of the first convolution layer of MobileNetV2 to 5 to match the input of BAENet, and change the output channels of the last fully connected layer to $n+1$ in order to predict the exposure times for $n$ burst images.
Here, the $(n+1)$-th element of the output vector allows the sum of the $n$ exposure times to be shorter than the upper bound, if necessary.

\section{Details of Noise Synthesis}
\label{sec:noise}

In the differentiable burst simulator, we synthesize realistic noise using a heteroscedastic Gaussian distribution, following \cite{BracketIRE}: 
\begin{equation}
    N \sim \mathcal{N}\left(0, \lambda_\textrm{read} + \lambda_\textrm{shot} S_{s,e} \right) \label{eq:noise},
\end{equation} 
where $\lambda_\textrm{shot}$ and $\lambda_\textrm{read}$ represent the shot noise and read noise parameters, respectively, and $S_{s,e}$ denotes the integration of scene radiance during the exposure duration, corresponding to Eq. (\textcolor{cvprblue}{7}) in the main paper.
The amount of noise should be proportional to the gain of each burst image $g_i$. 
In the following section, we describe how to determine the noise parameters according to $g_i$.
In this paper, we use the term \textquotedblleft gain\textquotedblright~to refer to the linear gain, which is equivalent to the camera ISO.

Given a preview image, we compute the gain of each burst image $g_i$ as:
\begin{equation}
    g_i = g_p \cdot \frac{t_p}{t_i},
\end{equation} 
where $t_i$ is the exposure time estimated by BAENet for each burst image, and $g_p$ and $t_p$ are the gain and exposure time of the preview image, respectively. 
To simulate extremely low-light images, we randomly sample $g_p$ from a uniform distribution within the range $[51200, 102400]$.

Based on $g_i$, we compute the shot noise parameter ${\lambda}^{i}_{shot}$ and the read noise parameter ${\lambda}^{i}_{read}$ for each burst image.
To this end, we first calibrated the shot noise and read noise parameters of our camera system across a range of gains, from 100 to 12,800.
Then, we model the relationship between gains and the shot and read noise parameters, following~\cite{brooks2019unprocessing}.
The shot noise parameter is linearly proportional to the gain.
For each burst image, we model the shot noise parameter ${\lambda}^{i}_{shot}$ as:
\begin{equation}
    \lambda^{i}_{shot} = 9.2857 \mathrm{e}{-07} \times g_i + 8.1006 \mathrm{e}{-05} \label{eq:shot_noise},
\end{equation} 
The coefficients of \cref{eq:shot_noise} are estimated from the calibrated shot noise parameters in our camera system.
Additionally, read noise parameters are known to be linearly proportional to the shot noise parameters in the log domain.
We model the read noise parameter $\lambda^{i}_{read}$ as:
\begin{equation}
    \log(\lambda^{i}_{read}) = 2.2282 \times \log(\lambda^{i}_{shot}) + 0.45982,  
    \label{eq:read_noise}
\end{equation} 
The coefficients of \cref{eq:read_noise} are also estimated from the calibrated read noise and shot noise parameters.
For each burst image, we compute $\lambda^{i}_{shot}$ and $\lambda^{i}_{read}$ according to $g_i$ and then synthesize noisy burst images using \cref{eq:noise}.

\section{Details of RAW conversion}
\label{sec:raw}

In our dataset, we convert sRGB video clips into RAW video clips. 
We first apply gamma expansion to convert the video clip into the linear sRGB space, then convert its color space to the RAW color space using color correction matrices (CCMs). 
Next, we apply inverse white balance and mosaic the resulting videos to the RGGB Bayer pattern.
In this section, we describe the details of color space conversion and inverse white balance.

\paragraph{Conversion to RAW Color Space}
Camera systems capture images in the RAW color space and convert them to sRGB images using color correction matrices (CCMs).
Using the inverse of CCMs, we can convert the sRGB videos in our dataset to the RAW color space as follows:
\begin{equation}
    \begin{bmatrix}
        R_{\text{RAW}} \\
        G_{\text{RAW}} \\
        B_{\text{RAW}}
    \end{bmatrix}
    =
    \text{CCM}^{-1}
    \cdot
    \begin{bmatrix}
        R_{\text{sRGB}} \\
        G_{\text{sRGB}} \\
        B_{\text{sRGB}}
    \end{bmatrix},
\end{equation}
where $(R_{\text{sRGB}}, G_{\text{sRGB}}, B_{\text{sRGB}})$ is an RGB color of the sRGB color space. $(R_{\text{RAW}}, G_{\text{RAW}}, B_{\text{RAW}})$ is the corresponding RGB color converted to the RAW color space. $\text{CCM}$ is a color correction matrix.
To obtain the matrix, we calibrated three $\text{CCMs}$ using the target camera (Basler a2A1920-160ucBAS) under three different scenes.
We randomly sample $\text{CCM}$ from the three estimated matrices for each video clip and convert the video clip to the RAW color space.
This reflects the fact that the $\text{CCM}$ of real camera systems can vary under different light sources.

\paragraph{Inverse White Balance}
For each video clip, we randomly sample white balance gains to reflect varying white balance adjustments in real-world scenarios. Then, we perform inverse white balance as follows:
\begin{equation}
    \small  
    R = R' \cdot \frac{g_{RGB}}{g_R}, \quad G = G' \cdot g_{RGB}, \quad B = B' \cdot \frac{g_{RGB}}{g_B} \label{eq:whitebalance},
\end{equation}
where $R', G', B'$ and $R, G, B$ are the pixel values of the RGB channels before and after inverse white balance, respectively. 
$g_{RGB}$ represents the total gain for all RGB channels, while $g_R$ and $g_B$ are the individual white balance gains for red and blue channels. 
The white balance gain for the green channel is typically set to one, so we omit it in \cref{eq:whitebalance}.
We randomly sample $g_{RGB}$ from a Gaussian distribution $\mathcal{N}\left(0.8, 0.1 \right)$. $g_R$ and $g_B$ are sampled from uniform distributions $\mathcal{U}\left(1.9, 2.4 \right)$ and $\mathcal{U}\left(1.5, 1.9 \right)$, respectively.

\section{Details of the Dataset}
\label{sec:dataset}

Along with scene radiance sequences with camera and object motion from the GoPro dataset~\cite{nah2017deep}, we include static sequences without such motion.
To achieve this, we utilize ground-truth sharp images from the RealBlur dataset~\cite{rim_2020_ECCV}. 
Specifically, we sample a sharp sRGB image from the RealBlur dataset and convert it into the camera RAW color space as described in Sec. \textcolor{cvprblue}{5.1} of the main paper.
We then generate a scene radiance sequence by duplicating it. 
Although assuming no camera or object motion in burst shots is less realistic, we found empirically that including static scenes stabilizes the training process and enhances the performance of our framework.
Finally, we generated $\mathcal{D}_\textrm{restore}$ and $\mathcal{D}_\textrm{BAENet}$ containing a total of 4,092 and 1,127 sequences, respectively, with 3,728 and 1,036 from the GoPro dataset and 364 and 91 from the RealBlur dataset.
For evaluation of our method, we also generated a test set consisting of 532 scene-radiance sequences from the GoPro dataset.

Since our synthetic training dataset uses frames from daytime high-FPS videos~\cite{nah2017deep} as the source images for burst simulation, a potential sim-to-real gap may arise.
As shown in Fig.~\textcolor{cvprblue}{2} of the main paper, the preview and input burst images are captured in daytime conditions, which differ from the extremely low-light scenarios targeted by our method.
Nevertheless, prior work~\cite{rim2022realistic} has shown that models trained on blurred images synthesized from daytime high-FPS videos with added noise can generalize well to real low-light data~\cite{rim_2020_ECCV}.
Consistent with these findings, our real-capture experiments presented in Sec.~\textcolor{cvprblue}{6.2} of the main paper demonstrate that the proposed method performs reliably in real-world low-light scenarios.
At the same time, the sim-to-real gap cannot be entirely eliminated and remains a natural limitation when training on synthetic data.
Addressing this gap is an important direction for future work.

\section{Camera System for Real-world Evaluation}
\label{sec:camera}

In the real-world capturing setup described in Sec. \textcolor{cvprblue}{6.2} of the main paper, our handheld dual-camera system consists of two Basler a2A1920-160ucBAS cameras, each equipped with an Edmund 6 mm C-Series lens.
One camera captures burst images with exposure times predicted by BAENet, while the other uses a predefined exposure setting of $\{8,24,40,56\}/1920$ seconds.
The system was connected to a Samsung Galaxy Book4 Ultra NT960XGP laptop, and we captured 142 burst images in low-light environments.

\section{Computational Cost}
\label{sec:cost}
The computational workload of our pipeline consists of three components: motion estimation from preview images using RAFT-small~\cite{teed2020raft}, BAENet, and burst image restoration. 
Other methods in Tab.~\textcolor{cvprblue}{1} rely on heuristic strategies or lightweight predictors for exposure selection, so their computational cost mainly arises from burst image restoration (1769G FLOPs).
In comparison, DEBIR additionally requires motion estimation and exposure prediction using BAENet, which together introduce only 71G FLOPs as reported in \cref{tab:comp-cost}. 
Therefore, the additional overhead of DEBIR is negligible compared to the overall cost of burst image restoration.

\begin{table}[t!]
    \centering
    \scalebox{0.80}{ 
    \begin{tabularx}{1.225\columnwidth}{cc@{\hspace{0.8em}}c@{\hspace{0.8em}}c} 
    \Xhline{2.5\arrayrulewidth} 
    Components & Motion estimation & BAENet & Burst image restoration \\ \hline
    FLOPs (G) & $65$ & $6$ & $1769$ \\  
    \Xhline{2.5\arrayrulewidth}
    \end{tabularx}
    }
    \vspace{-0.2cm} 
    \caption{Computational cost of each module in our pipeline.}
    \vspace{-0.2cm}
    \label{tab:comp-cost}
\end{table}

\begin{figure}[t]
\includegraphics[width=1.0\linewidth]{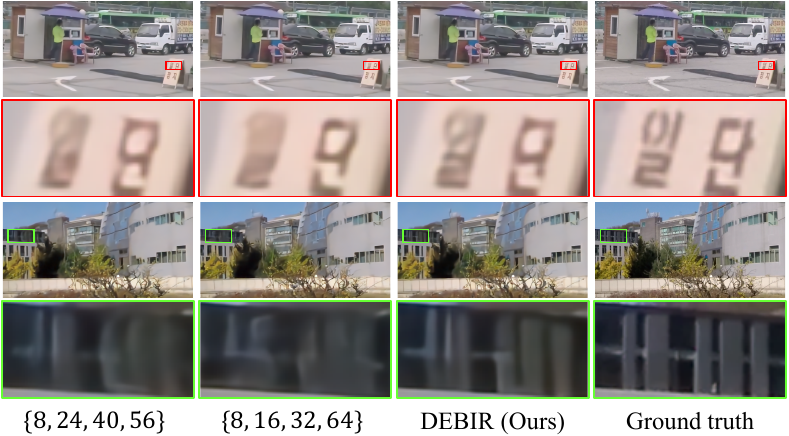}
\centering
\vspace{-0.6cm}
\caption{Qualitative comparison with another predefined-exposure schedule on our test set.}
\vspace{-0.2cm}
\label{fig:pre_comb}
\end{figure}

\begin{table}[t]
    \centering
    \scalebox{0.88}{
    \begin{tabularx}{1.07\columnwidth}{c|ccc}
    \Xhline{2.5\arrayrulewidth}
    Metric & $\{8,24,40,56\}$ & $\{8,16,32,64\}$ & DEBIR (Ours) \\
    \hline
    PSNR$\uparrow$ & 35.04 & \snd{35.07} & \fst{35.32} \\
    SSIM$\uparrow$ & 0.9481 & \snd{0.9482} & \fst{0.9519} \\
    LPIPS$\downarrow$ & 0.164 & \snd{0.163} & \fst{0.154} \\
    \Xhline{2.5\arrayrulewidth}
    \end{tabularx}
    }
    \vspace{-0.2cm}
    \caption{Quantitative comparison with another predefined-exposure schedule on our test set. The best and second-best results are in bold and blue, respectively.}
    \vspace{-0.4cm}
    \label{tab:pre_comb}
\end{table}

\section{Additional Experiments}
\label{sec:experiments-supp}

\paragraph{Predefined Exposure Schedule}
In the main paper, we compared against a predefined-exposure baseline~\cite{kim2024burst,BracketIRE} based on an arithmetic schedule, $\{8, 24, 40, 56\}/1920$ seconds, where the exposure times increase by a fixed step of $16/1920$ seconds. 
Since HDR pipelines often employ exponentially increasing exposure times, we additionally evaluated a geometric schedule, $\{8, 16, 32, 64\}/1920$ seconds, which follows a constant ratio of $2$. 
As shown in \cref{fig:pre_comb}, DEBIR produces finer details than the predefined schedules.
Consistently, \cref{tab:pre_comb} shows that both predefined schedules yield similar but overall inferior performance, whereas our DEBIR model achieves the best results across all metrics.
This is because predefined exposure bracketing follows a fixed schedule that cannot account for varying shooting environments, whereas DEBIR can adapt to the scene and thus achieves better results.

\paragraph{Performance Gap across Noise Levels}
To further analyze the results in Tab.~\textcolor{cvprblue}{1} of the main paper, we evaluate performance across different noise levels.
Our test set is synthesized with noise corresponding to ISO values sampled from the range [51200, 102400].
We divide the test set into five bins with evenly spaced ISO intervals and evaluate performance in each bin, as summarized in \cref{tab:noise_level}.

\begin{figure}[t]
\includegraphics[width=1.0\linewidth]{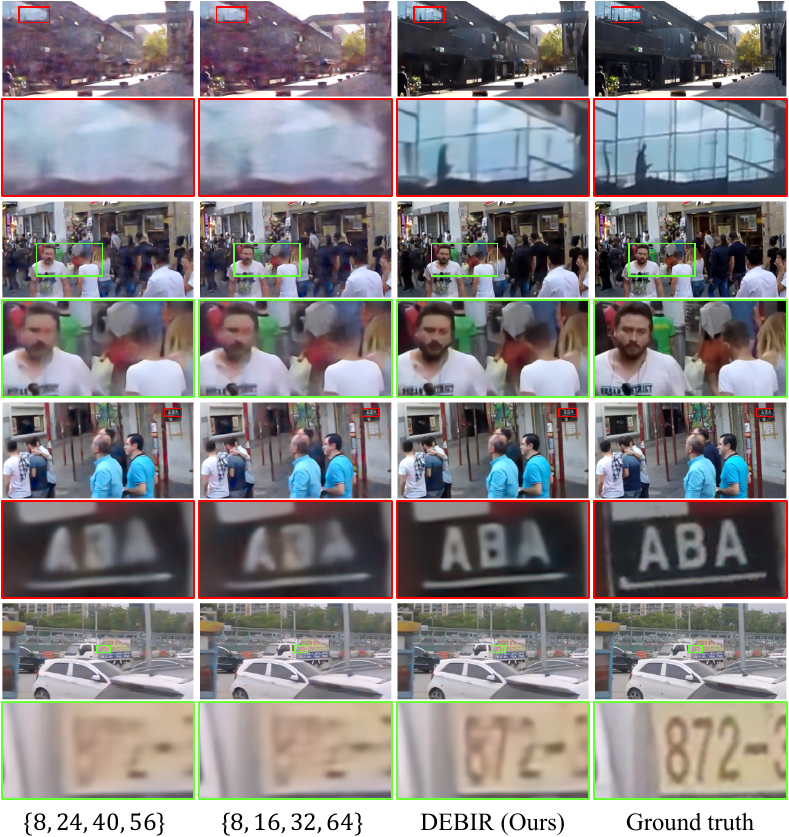}
\centering
\vspace{-0.6cm}
\caption{Qualitative comparison with Exposure Bracket~\cite{kim2024burst,BracketIRE} on the new test set with stronger noise.}
\vspace{-0.2cm}
\label{fig:noise_level}
\end{figure}

\begin{table}[t]
    \centering
    \scalebox{0.88}{
    \begin{tabularx}{1.13\columnwidth}{c|c@{\hspace{0.9em}}c@{\hspace{0.9em}}c@{\hspace{0.9em}}c@{\hspace{0.9em}}c|c}
    \Xhline{2.5\arrayrulewidth}
    \multirow{2}{*}{Methods} & \multicolumn{5}{c|}{Original (Tab.~\textcolor{cvprblue}{1})} & \multirow{2}{*}{New} \\
    \cline{2-6}
     & 1 & 2 & 3 & 4 & 5 &  \\ 
    \hline
    $\{8,24,40,56\}$ & 35.75 & 35.40 & 35.15 & 34.35 & 34.70 & \snd{32.04} \\
    $\{8,16,32,64\}$ & \snd{35.81} & \snd{35.43} & \snd{35.20} & \snd{34.36} & \snd{34.71} & 32.00 \\
    DEBIR (Ours) & \fst{35.97} & \fst{35.64} & \fst{35.43} & \fst{34.65} & \fst{35.03} & \fst{32.87} \\
    \Xhline{2.5\arrayrulewidth}
    \end{tabularx}
    }
    \vspace{-0.2cm}
    \caption{PSNR comparison with Exposure Bracket. 
    Columns 1–5 correspond to ISO-based subsets of our test set, while column 6 corresponds to a new test set with stronger noise.
    The best and second-best results are shown in bold and blue, respectively.}
    \vspace{-0.1cm}
    \label{tab:noise_level}
\end{table}

The first two rows in \cref{tab:noise_level} correspond to the predefined-exposure baseline Exposure Bracket~\cite{kim2024burst,BracketIRE} using arithmetic and geometric exposure schedules $\{8, 24, 40, 56\}/1920$ and $\{8, 16, 32, 64\}/1920$ seconds, respectively.
Across all noise levels (columns 1–5 in \cref{tab:noise_level}, with increasing ISO values), DEBIR consistently achieves higher PSNR than Exposure Bracket, and the performance gap becomes larger as the ISO value increases.
We further construct a new test set using the same scenes with ISO values sampled from a higher range [153600, 204800], resulting in significantly stronger noise.
The corresponding result is reported in column 6 (New) of \cref{tab:noise_level}, where the performance gap between Exposure Bracket and DEBIR increases to 0.83 dB.

Qualitative results in \cref{fig:noise_level} also support this observation.
DEBIR better preserves fine details while producing fewer color artifacts caused by denoising failures.
Overall, these results demonstrate that DEBIR shows a clear advantage under harsher low-light conditions with stronger noise.
By adaptively predicting exposure times according to the shooting environment, DEBIR enables more effective burst image restoration.

\begin{table}[t]
    \centering
    \scalebox{0.88}{
    \begin{tabularx}{1.13\columnwidth}  {c|c@{\hspace{0.65em}}c@{\hspace{0.65em}}c@{\hspace{0.65em}}c@{\hspace{0.65em}}c@{\hspace{0.65em}}c}
    \Xhline{2.5\arrayrulewidth}
    \multirow{3}{*}{Test set} & \multirow{3}{*}{\shortstack{Dig.\vspace{-0.07cm}\\Gim.\\ \cite{dahary2021digital}}} & \multirow{3}{*}{\shortstack{Act.\\S-L\\ \cite{yang2022active}}} & \multirow{3}{*}{\shortstack{Avg.\vspace{-0.07cm}\\AE\\ \cite{cameraProduct}}} & \multirow{3}{*}{\shortstack{Grad.\\AE\\ \cite{shim2018gradient}}} & \multirow{3}{*}{\shortstack{Exp.\vspace{-0.07cm}\\Brk.\\ \cite{kim2024burst,BracketIRE}}} & \multirow{3}{*}{\shortstack{DEBIR\\(Ours)}} \\
    \\ \\ \hline
    REDS $+$ Basler & 32.53 & 32.44 & 33.44 & \snd{33.68} & 33.60 & \fst{33.86} \\
    REDS $+$ Pixel & 31.44 & 31.26 & 32.29 & 32.44 & \snd{32.45} & \fst{32.50} \\  
    \Xhline{2.5\arrayrulewidth}
    \end{tabularx}
    }
    \vspace{-0.2cm}
    \caption{Quantitative comparison of PSNR on new test sets. The best and second-best results are in bold and blue, respectively.}
    \vspace{-0.4cm}
    \label{tab:val-on-reds}
\end{table}

\paragraph{Generalization Capability}
In image restoration tasks, neural networks trained on data from a specific dataset or camera are not necessarily expected to generalize well to different datasets or devices~\cite{rim_2020_ECCV}. 
Our setup follows common practice in image restoration, where models are typically trained for a target camera using datasets tailored to that device.
However, it may still be questioned whether our model achieves strong performance only on the datasets or the noise characteristics of the target camera used in our experiments, i.e., whether the reported performance is simply a result of overfitting to the datasets used.
To this end, we conducted an additional study on generalizability. 

Specifically, we generated two test sets using the REDS dataset~\cite{Nah_2019_CVPR_Workshops_REDS} with noise parameters from two different cameras: a Basler camera, which was used in our main experiments, and a Google Pixel. 
For the Google Pixel, we used the calibrated noise parameters provided in the SIDD dataset~\cite{SIDD_2018_CVPR}. 
As our training set is generated using sharp frames from the GoPro~\cite{nah2017deep} and RealBlur~\cite{rim_2020_ECCV} datasets and the noise characteristics of a Basler camera, these test sets allow us to examine generalization to unseen scenes captured by both the same and a different device.

As shown in \cref{tab:val-on-reds}, our method outperforms other approaches on `REDS + Basler', demonstrating strong generalization to novel scenes from the same camera. 
Furthermore, while cross-device generalization is inherently challenging, our results on `REDS + Pixel' suggest that the model may still outperform competing methods even on images captured by a different camera.
Furthermore, since our pipeline synthesizes training data from camera parameters, adapting the framework to a different target camera is straightforward in practice.

\paragraph{The Impact of Each Burst Image}
In this study, we examine which image among the burst images is most important in our pipeline.
To verify this, we replace each burst image with a zero-filled image and evaluate the performance.
\cref{tab:abl-bursts} shows that replacing any burst image significantly reduces performance, indicating that all burst images play a crucial role in restoration.
Notably, the first burst image, which is the base frame in the burst image restoration network, has the most significant impact on performance.
The influence gradually decreases for subsequent frames.

\begin{table}[t]
    \centering
    \scalebox{0.88}{
    \begin{tabularx}{1.09\columnwidth}{cc@{\hspace{0.86em}}c@{\hspace{0.86em}}c@{\hspace{0.86em}}c@{\hspace{0.86em}}c}
    \Xhline{2.5\arrayrulewidth}
    $i$-th burst image & w/o $I_1$ & w/o $I_2$ & w/o $I_3$ & w/o $I_4$ & Full \\ \hline
    PSNR$\uparrow$ & 13.71 & 32.12 & 33.91 & \snd{34.89} & \fst{35.32} \\
    SSIM$\uparrow$ & 0.1059 & 0.9161 & 0.9367 & \snd{0.9461} & \fst{0.9519} \\
    LPIPS$\downarrow$ & 0.627 & 0.198 & 0.183 & \snd{0.167} & \fst{0.154} \\ 
    \Xhline{2.5\arrayrulewidth}
    \end{tabularx}
    }
    \vspace{-0.2cm}
    \caption{Analysis of the impact of each burst image on performance. Here, $I_i$ refers to the $i$-th burst image. The best and second-best results are in bold and blue, respectively.}
    \vspace{-0.4cm}
    \label{tab:abl-bursts}
\end{table}

\paragraph{Comparison with RL-based Methods}
Several RL-based methods~\cite{xu2025adaptiveae,wang2020learning} have been proposed for automatic exposure control. 
However, most of them target HDR reconstruction and are not well suited for burst image restoration. 
AdaptiveAE~\cite{xu2025adaptiveae} predicts exposure times in a step-wise manner, resulting in a sequential decision-and-capture pipeline that is inefficient and unsuitable for burst image restoration scenarios.
Another RL-based method~\cite{wang2020learning} assumes static scenes and operates on a small, fixed number of exposures (e.g., three), which makes it difficult to extend methodologically to burst image restoration settings that typically require more frames.

Nevertheless, we evaluate a publicly available RL-based method~\cite{wang2020learning} for comparison.
To ensure a fair comparison, we fine-tuned Burstormer~\cite{dudhane2023burstormer} using exposure times predicted by the policy network of~\cite{wang2020learning}.
The corresponding burst images were synthesized using our burst simulator.
Using this approach, the method achieved 34.71 dB PSNR, which is 0.61 dB lower than DEBIR.
While our focus is burst image restoration, extending DEBIR to support HDR reconstruction is an interesting direction for future work.

\paragraph{Motion Estimation Reliability}
In our implementation, preview images are captured with an exposure time of $1/120$ seconds, which we empirically found to provide sufficiently sharp previews for reliable flow estimation, whereas longer exposure times introduce motion blur.
Under this setting, we further evaluated the robustness of motion estimation by estimating the motion magnitude $m_p$ using noise-free preview images and observed only a negligible performance difference of 0.04 dB. 
This robustness stems from how the motion cue $m_p$ is defined in our pipeline: optical flow is computed on downsampled preview images and aggregated into a single scalar by averaging over the entire frame, intentionally capturing only coarse motion information. 
Although optical flow estimation may degrade under more severe conditions, the estimator itself can be further fine-tuned to improve robustness if necessary. 
Moreover, our framework is not restricted to optical flow. Alternative motion cues such as gyroscope signals can also be incorporated without modifying the pipeline.

\section{Additional Results}
\label{sec:results-supp}

\begin{figure*}[p]
\centering
\includegraphics[width=1\linewidth]{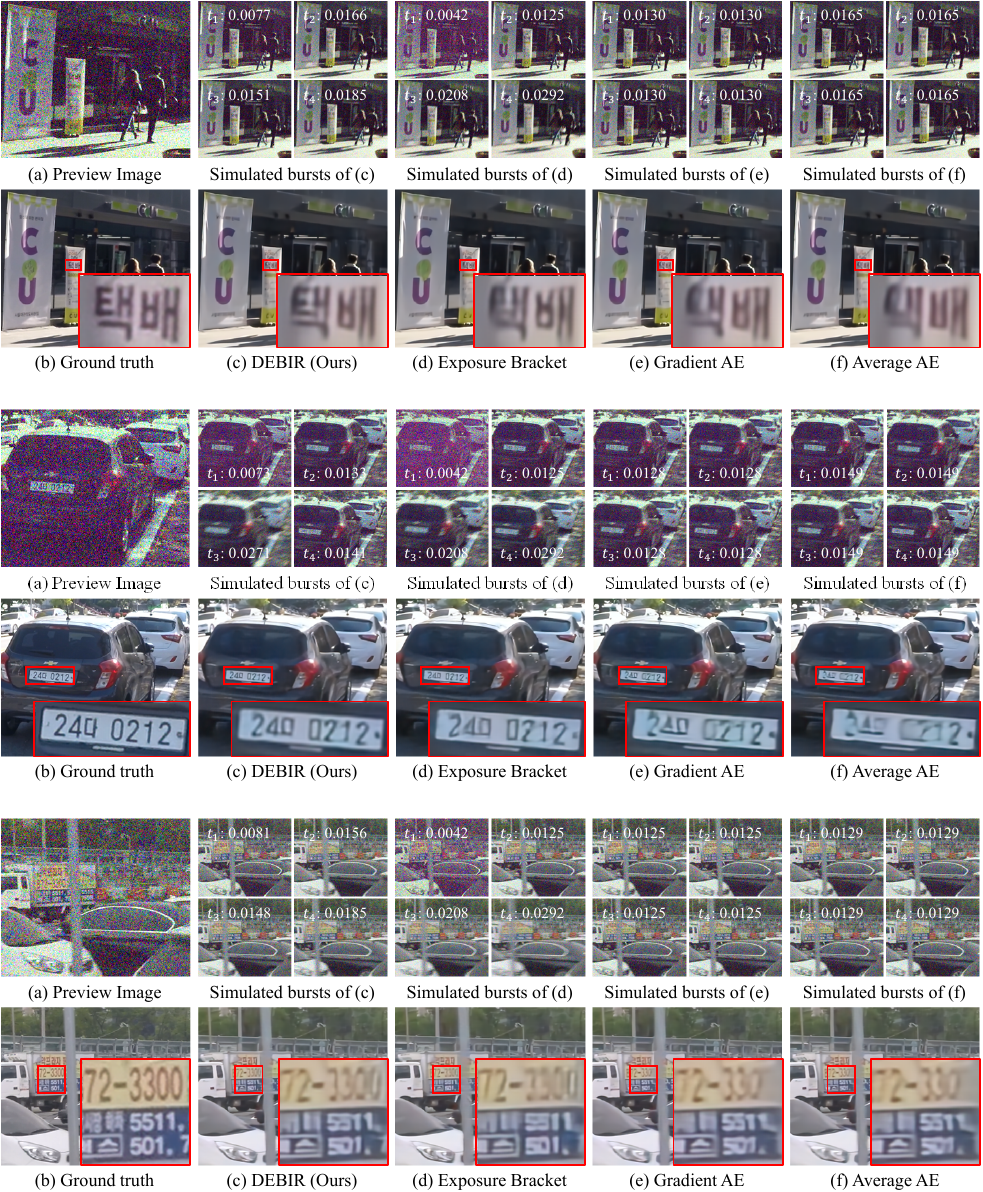}
\vspace{-0.2cm}
\caption{Additional qualitative results on our test set.}
\label{fig:supp-syn}
\end{figure*}

\begin{figure*}[p]
\centering
\includegraphics[width=0.96\linewidth]{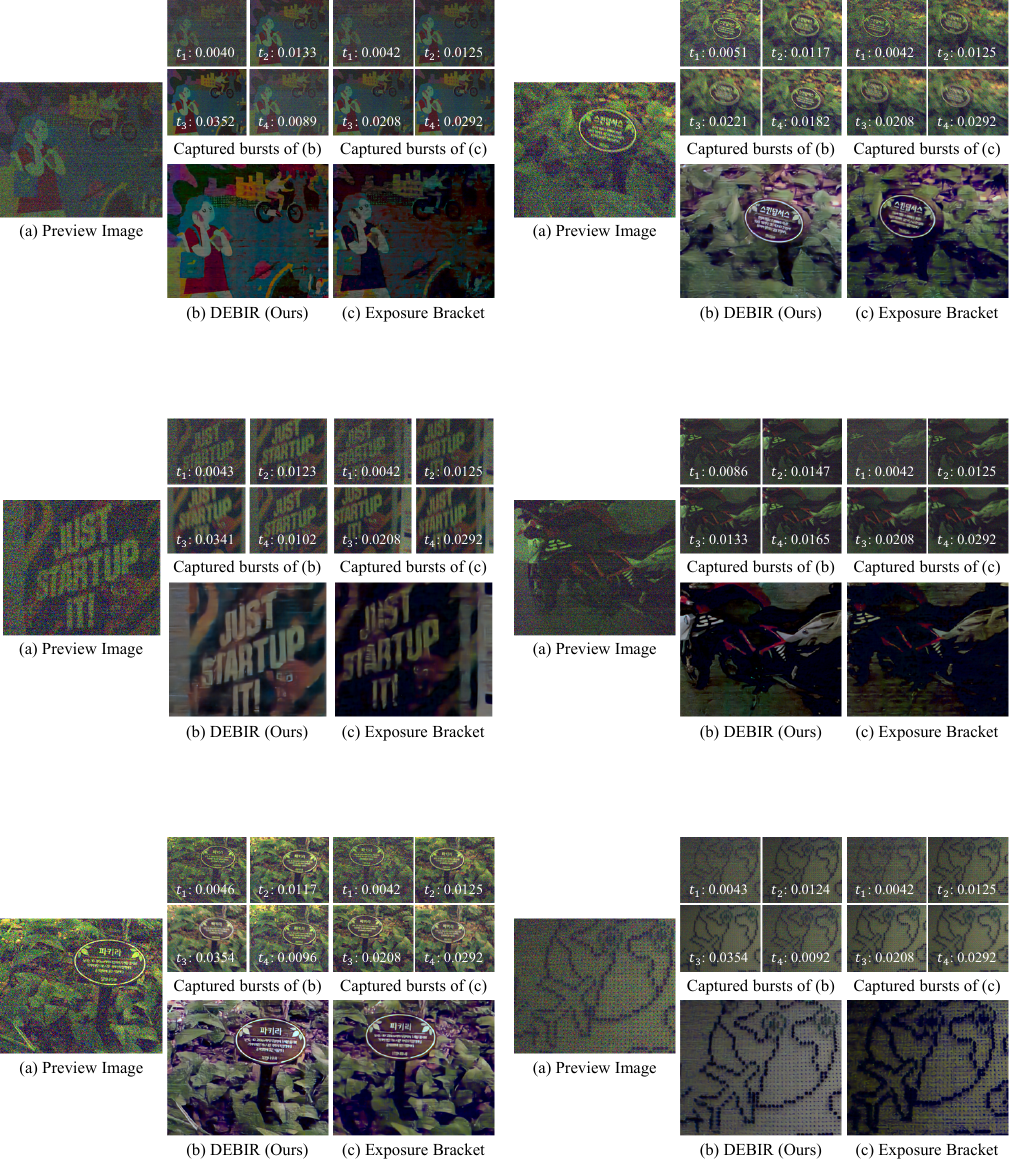}
\caption{Additional qualitative results using a real-world camera system.}
\label{fig:supp-real}
\end{figure*}

We present additional qualitative results on our test set (\cref{fig:supp-syn}) and real-world results obtained with our camera system (\cref{fig:supp-real}). 
These results demonstrate that DEBIR yields visually superior outcomes over other methods, confirming its effectiveness.

{
    \small
    \bibliographystyle{ieeenat_fullname}
    \bibliography{references_supp}
}